\documentclass[letterpaper, conference]{ieeeconf}
\IEEEoverridecommandlockouts   
\usepackage{amsmath,amsfonts}
\usepackage{algorithmic}
\usepackage{algorithm}
\usepackage{array}
\usepackage[caption=false,font=normalsize,labelfont=sf,textfont=sf]{subfig}
\usepackage{textcomp}
\usepackage{stfloats}
\usepackage{url}
\usepackage{verbatim}
\usepackage{graphicx}
\usepackage{cite}
\usepackage[colorlinks,linkcolor=blue]{hyperref}
\usepackage{mathrsfs}
\usepackage[flushleft]{threeparttable}
\usepackage{tablefootnote}
\usepackage{multirow}
\usepackage{graphicx}
\usepackage{color}
\usepackage{tikz}
\usetikzlibrary{calc,arrows,decorations.markings}
\usepackage{balance}
\usepackage{booktabs}
\usepackage{xpatch}
\usepackage{mathrsfs}
\usepackage{float}

\makeatletter
\usepackage{etoolbox,lipsum}
\patchcmd\@makecaption{\\}{.~}{}{\fail}

\graphicspath{{./imgs/}}

\usepackage{caption}

\usepackage[most]{tcolorbox}

  \definecolor{findingborder}{RGB}{55,105,155}
  \definecolor{findingbackground}{RGB}{237,244,250}

  \newtcolorbox{findingbox}[1]{
      colback=findingbackground,
      colframe=findingborder,
      title={#1},
      fonttitle=\bfseries,
      rounded corners,
  }

\title{\LARGE \bf What Matters in Designing World Action Models: \\ An Empirical Study}

\author{Anonymous Authors}

  \author{
    Chao Tang\textsuperscript{1,*},
    Haoqing Wang\textsuperscript{2,*},
    Zilang Cen\textsuperscript{1,3,4},
    Weishi Mi\textsuperscript{1},
    Wei Xia\textsuperscript{5},
    Fangcheng Liu\textsuperscript{2},
    Anda Cheng\textsuperscript{2}, \\
    Yeqing Shen\textsuperscript{2},
    Xiaohui Cui\textsuperscript{4},
    Xiaoyuan Zhang\textsuperscript{3},
    Yehui Tang\textsuperscript{2},
    and Tingguang Li\textsuperscript{1} \\[0.5em]
    \small
  \textsuperscript{1}Samsung Robotics eXperience
  \quad
  \textsuperscript{2}Samsung R\&D Institute China--Beijing
  \quad
  \textsuperscript{3}Zhongguancun Academy \\
  \textsuperscript{4}Wuhan University
  \quad
  \textsuperscript{5}Peking University
  }

\begin{document}

\maketitle
\thispagestyle{empty}
\pagestyle{empty}

  \begingroup
  \renewcommand{\thefootnote}{*}
  \footnotetext{These authors contributed equally.}
  \endgroup

\begin{abstract}
World Action Models (WAMs) have emerged as a promising paradigm for generalizable robot control. Despite the growing number of WAM systems, existing works often introduce unified systems that bundle together multiple design choices, such as architecture and training strategy, making it difficult to isolate individual contributions and systematically compare alternative designs. In this work, we present a controlled study that disentangles these design choices and analyzes not only their empirical effects, but also how and why they shape WAMs. More specifically, we focus on three fundamental questions in building WAMs: (1) what causal structure should govern the interaction between world modeling and action generation? (2) in which latent space should world modeling be performed? and (3) how do different world-action modeling objectives affect model behavior and performance? Through structurally controlled experiments on three representative benchmarks, RoboCasa-GR1, LIBERO, and LIBERO-Plus, we systematically compare six causal structures, eight latent representations, and four training objectives, covering popular design choices in existing WAMs. We further validate our key findings on real-robot data from the DROID dataset. We hope to provide a systematic understanding of how core design choices affect world-action modeling and what principles can guide the development of future WAM systems.
\end{abstract}

% by jointly modeling physical dynamics and action generation

 % why certain designs work better than others, 

 % Note that the goal is not to build a WAM with the highest success rates on public benchmarks; rather, 

% \begin{IEEEkeywords}
% Robotic Grasping, Perception for Grasping and Manipulation, Deep Learning in Grasping and Manipulation
% \end{IEEEkeywords}

\section{Introduction}

Building generalizable robot policies capable of perceiving, reasoning, and interacting in the real world remains a central challenge. Recently, World Action Models (WAMs) \cite{wang2026world, hou2026world} have emerged as a promising paradigm for this goal. Despite the growing number of WAM systems \cite{ye2026world, yuan2026fast, pai2025mimic, ma2026dit4dit, ye2026gigaworld, kim2026cosmos, li2026causal, lyu2026lda}, relatively limited effort has been devoted to systematically understanding what matters most across different dimensions of WAM design (e.g., model structure, representation space). 

Existing studies either evaluate complete systems as a whole, where multiple components are introduced and optimized together, or examine only a narrow subset of design dimensions, leaving the broader WAM design space insufficiently understood. This makes it difficult to tell whether the observed gains come from better world representations, more effective world-action conditioning, more informative auxiliary objectives, or simply a larger and more carefully engineered system. As a result, the field lacks a controlled study that separates these factors and explains how each design dimension shapes both policy behavior and learned world-action representations.

% Instead of directly mapping observations to actions as in Vision-Language-Action (VLA) models, WAMs introduce world-modeling objectives into policy learning, encouraging the model to capture physical dynamics while generating actions. 

% Such designs are appealing as world-action modeling can provide richer supervision for learning how physical states evolve over time and respond to actions, which is essential for robust and generalizable robot control.

\begin{figure*}[t]
  \centering
  % \vspace*{-0.2in}
  \begin{tikzpicture}[inner sep = 0pt, outer sep = 0pt]
    \node[anchor=south west] (fnC) at (0in,0in)
      {\includegraphics[height=2.2in,clip=true,trim=0in 0in 0in 0in]{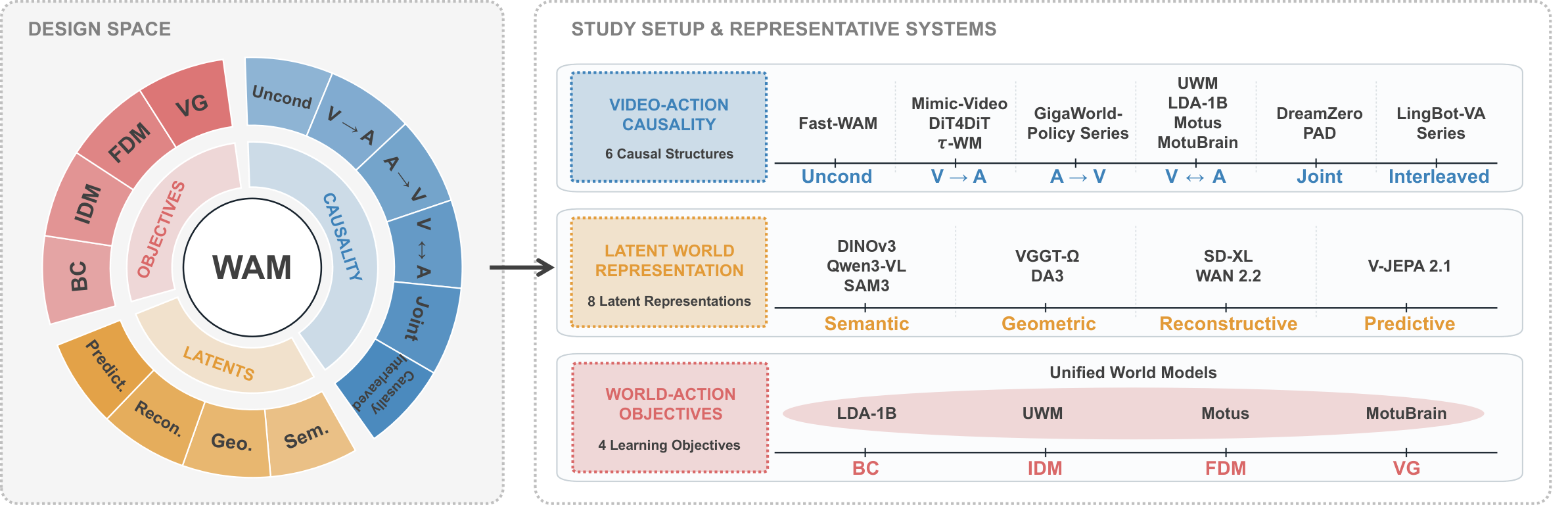}};
  \end{tikzpicture}
        \caption{This work systematically investigates three fundamental aspects in building WAMs: (1) video-action causality, (2) latent world representation, and (3) world-action modeling objectives.}
  \label{fig:teaser}
  % \vspace*{-0.2in}
\end{figure*}

In this work, we aim to answer these questions through a controlled empirical study. Rather than proposing a single new architecture optimized for benchmark success rates, we disentangle several core design choices and analyze how they shape WAM behavior and performance. As illustrated in Figure \ref{fig:teaser}, we focus on three fundamental questions. First, we study the \textbf{causal structure} between world modeling, typically instantiated as video generation, and action generation by comparing six representative video-action dependency patterns: disentangled/unconditional \cite{yuan2026fast}, video-to-action \cite{pai2025mimic, ma2026dit4dit, zhou2606tau0}, action-to-video \cite{ye2026gigaworld, team2026gigaworld}, bidirectional \cite{zhu2025unified, guo2026unified, lyu2026lda, bi2026motus, team2026motubrain}, joint \cite{ye2026world, guo2024prediction}, and causally interleaved \cite{li2026causal, zhang2026native} generation. Second, we analyze how the choice of \textbf{latent representation} shapes world modeling across four families: semantic \cite{simeoni2025dinov3, bai2025qwen3, carion2026sam}, geometric \cite{wang2026vggt, lin2025depth}, reconstructive \cite{podell2024sdxl, wan2025wan}, and predictive \cite{mur2026v} latents. Finally, we investigate how four \textbf{world–action modeling objectives}, including behavior cloning, inverse dynamics, forward dynamics, and video generation, jointly influence model capabilities.

Within the scope of our experimental setup, structurally controlled experiments across three representative benchmarks, RoboCasa-GR1, LIBERO, and LIBERO-Plus, under both in-distribution (ID) and out-of-distribution (OOD) settings support the following conclusions:\par

\noindent\textbullet\ \textbf{Video-Action Causality:} Generated futures causally influence action generation primarily through their temporal organization, whereas policies largely tolerate corruption of the exact future content. \par
\noindent\textbullet\ \textbf{Latent World Representation:} Temporal structure in latent representations strongly shapes performance across distributions: inter-frame latents favor ID control, whereas framewise latents provide greater OOD robustness.  \par
\noindent\textbullet\ \textbf{World-Action Modeling Objectives:} Behavior cloning and video generation provide relatively stable supervision under distribution shifts, whereas dynamics objectives may offer only marginal improvements or even cause negative transfer in some cases; in ID setting, however, auxiliary world modeling objectives generally degrade performance.

In addition to simulation experiments, we further validate key findings on real-robot data from the DROID dataset~\cite{khazatsky2024droid}.  In what follows, we present supporting evidence and analyses from these three perspectives, which we hope will offer useful insights into the development of future WAMs.

\section{Related Work}\label{related}

\subsection{World Action Models}

% WAM extends the VLA formulation by jointly predicting robot actions and future world states or representations, thereby integrating physical dynamics modeling into policy learning. 

Existing WAMs differ substantially along three dimensions: (1) the information flow between world modeling and action generation, (2) the latent space in which world dynamics are modeled, and (3) the auxiliary world–action objectives used alongside behavior cloning. For the first dimension, representative designs include disentangled/unconditional video and action prediction \cite{yuan2026fast}, video-to-action conditioning \cite{pai2025mimic, ma2026dit4dit, zhou2606tau0, li2026wall}, action-conditioned video prediction \cite{ye2026gigaworld, team2026gigaworld}, bidirectional interaction \cite{zhu2025unified, guo2026unified, lyu2026lda, bi2026motus, team2026motubrain, kim2026cosmos}, joint prediction \cite{guo2024prediction, ye2026world}, and causally interleaved generation \cite{li2026causal, zhang2026native}. These structures impose different assumptions about whether predicted futures should guide actions, actions should explain future states, or both modalities should be modeled symmetrically. Note that we classify WAMs by the defining video–action dependency of their canonical policy path. For instance, DreamZero is Joint because it co-denoises both modalities within each chunk, whereas LingBot-VA series is Causally Interleaved because it temporally orders them in a unified causal sequence; autoregressive chunk scheduling alone does not determine the category.

Another key distinction lies in the latent space used for world modeling. A broad class of WAMs \cite{yuan2026fast, ye2026gigaworld, team2026gigaworld, kim2026cosmos} operates in VAE latent spaces inherited from pretrained video generation models \cite{agarwal2025cosmos, wan2025wan}. In contrast, recent studies \cite{lyu2026lda, zhou2025dino} claim that pretrained DINO patch embeddings \cite{simeoni2025dinov3} capture rich semantic and object-centric information, making them particularly well suited for embodied world modeling. Extending this perspective, V-JEPA 2.1 \cite{mur2026v} argues that its latent representations combine fine-grained spatial grounding with temporal dynamics and global semantics, thereby further improving policy performance. Inspired by these arguments, we systematically analyze how the choice of latent representation shapes world modeling across four families: semantic \cite{simeoni2025dinov3, bai2025qwen3, carion2026sam}, geometric \cite{wang2026vggt, lin2025depth}, reconstructive \cite{podell2024sdxl, wan2025wan}, and predictive \cite{mur2026v} latents. A final dimension concerns recent unified world–action models \cite{lyu2026lda, zhu2025unified, bi2026motus, team2026motubrain}, which combine a conventional VLA-style behavior cloning (BC) objective with auxiliary world-modeling objectives, including forward dynamics modeling (FDM), inverse dynamics modeling (IDM), and video generation (VG). We therefore investigate how these four objectives influence model capabilities and empirical outcomes.

\subsection{Systematic Studies of Generalist Policy}

In recent years, several studies have systematically examined the design and evaluation of generalist robot policies. RoboVLMs \cite{li2026matters} investigates VLM backbones, policy architectures, history integration, and cross-embodiment
training, highlighting the importance of strong vision-language representations, effective history fusion, and in-domain data. OpenVLA-OFT \cite{kim2016fine} instead studies adaptation choices, showing that parallel decoding, action chunking, continuous action representations, and $\ell_1$ regression jointly improve policy success and inference efficiency. Related studies such as Octo \cite{team2024octo} and BAKU \cite{haldar2024baku} further analyze the effects of policy architecture, training data, observation history, and action heads. Beyond architectural design, robustness benchmarks \cite{guruprasad2024benchmarking, fei2026libero} reveal substantial sensitivity to changes in viewpoint, robot initialization, language, and scene appearance. A recent comparison of WAMs and VLAs \cite{zhang2026world} finds that WAMs are generally more robust to noise, illumination, and layout perturbations but remain vulnerable to geometric shifts and incur greater inference latency. Unlike these studies, which either focus on conventional VLAs or compare independently developed systems with multiple confounding differences, we conduct controlled studies by holding the base model fixed within each design dimension.

\section{Study Setup}

\subsection{Evaluation Benchmarks}

% \textbf{Benchmarks.} 

\textbf{Simulation Evaluation.} To systematically evaluate the above design choices, we benchmark all models on three representative simulation benchmarks. Specifically, RoboCasa-GR1 is used for ID evaluation, while LIBERO and LIBERO-Plus serve as OOD testbeds. RoboCasa-GR1 is a tabletop manipulation benchmark built on the RoboCasa simulation framework, in which a GR-1 humanoid robot performs bimanual manipulation with two dexterous hands. It comprises 24 language-conditioned tasks and 24K human-teleoperated demonstrations. LIBERO-Plus extends the standard LIBERO benchmark with controlled distribution shifts in object layouts, camera viewpoints, robot initial states, language instructions, lighting, background textures, and sensor noise. We evaluate models trained on LIBERO under these perturbations to measure OOD generalization. To isolate the effect of the training objectives, all compared policy variants are trained independently under matched settings and are not initialized from any previously trained policy checkpoint.

% spanning object rearrangement and articulated-object manipulation

\textbf{Real-Robot Data Evaluation.} We evaluate on real-robot data from DROID~\cite{khazatsky2024droid} by splitting the dataset into train/val/test sets based on task semantics. The objective is to further validate whether our key findings from the simulation generalize to real-robot data. All policy variants are trained independently under matched settings and evaluated via offline action prediction on the test split. Since binary success rates may not reveal fine-grained differences among policies in the real world, we report element-wise metrics in the normalized action space: mean squared error (MSE), mean absolute error (L1), Accuracy@0.1, and Accuracy@0.5, where Accuracy@$\tau$ denotes the fraction of valid action elements with absolute error at most $\tau$.

% \textbf{Evaluation Metrics.} \ Sucess rate as the main evaluation metric + auxiliary metric such as gradient cosine similarity and gradient norm. \textcolor{red}{NEED MORE DETAILS HERE}.

\begin{figure}[t]
  \centering
  % \vspace*{-0.2in}
  \begin{tikzpicture}[inner sep = 0pt, outer sep = 0pt]
    \node[anchor=south west] (fnC) at (0in,0in)
      {\includegraphics[height=2.5in,clip=true,trim=0.1in 0in 0in 0in]{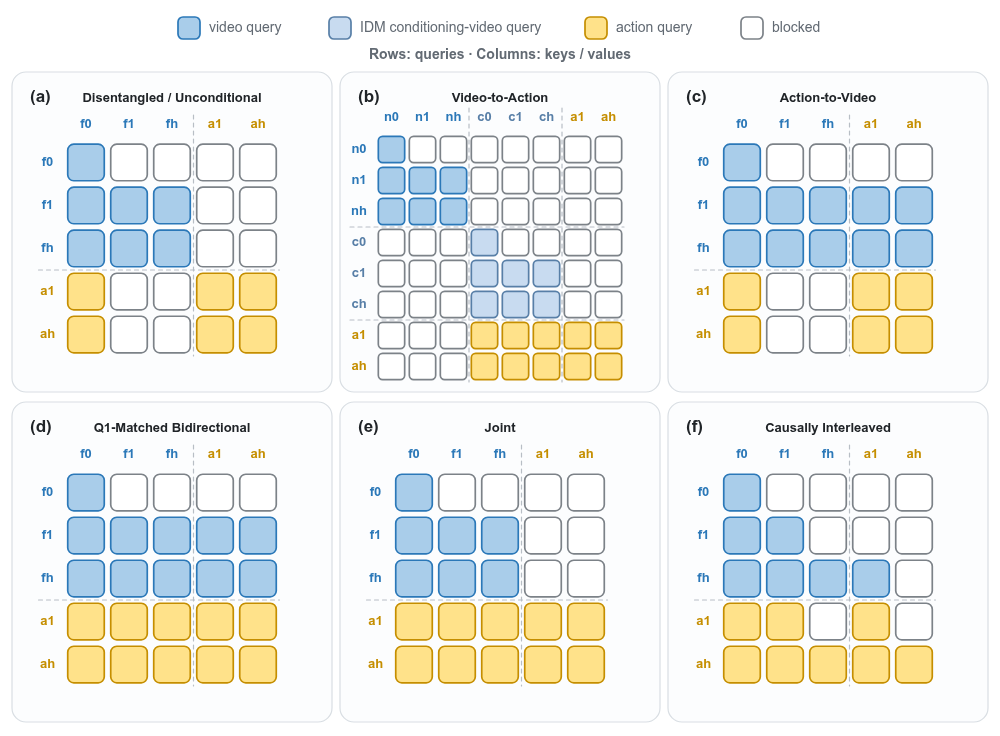}};
  \end{tikzpicture}
        \caption{Six representative video-action causal structures: (a) Disentangled/Unconditional,  (b)Video-to-Action, (c) Action-to-Video, (d) Bidirectional, (e) Joint, (f) Causally Interleaved.}
  \label{fig:attetion_mask}
  % \vspace*{-0.2in}
\end{figure} 

  % (a) Disentangled/Unconditional — FastWAM-Uncond
  % (b) Video-to-Action — FastWAM-IDM
  % (c) Action-to-Video — FastWAM-Action-to-Video
  % (d) Bidirectional — FastWAM-Bidirectional
  % (e) Joint — FastWAM-Joint
  % (f) Causally Interleaved — FastWAM-LingBot-Causal

\subsection{WAM Design Variants} \label{ss_variants}

\textbf{Video-Action Causality.}  We investigate six representative causal structures: disentangled/unconditional video and action prediction, video-conditioned action prediction, action-conditioned video prediction, bidirectional interaction, joint prediction, and causally interleaved generation. We instantiate and compare all structures within the same framework \cite{yuan2026fast}, while keeping the model architecture and training pipeline fixed. This controlled study reveals whether the benefits of world-action modeling arise from using predicted futures to guide actions, grounding future prediction in actions, jointly modeling both modalities, or simply retaining video prediction as auxiliary supervision. For the causally interleaved mode, we adopt a single-stream, group-causal abstraction that preserves the temporal video-action dependencies of \cite{li2026causal,zhang2026native} while ensuring architectural and training consistency across variants. Figure \ref{fig:attetion_mask} illustrates the evaluated causal structures.

% The direction of information flow between video and action tokens determines how a WAM couples dynamics learning with policy prediction.

\textbf{Latent World Representation.}  We investigate four complementary families: semantic latents, such as DINOv3~\cite{simeoni2025dinov3}, Qwen3-VL~\cite{bai2025qwen3}, and SAM3~\cite{carion2026sam} features, which emphasize semantic and object-centric information; geometric latents, which encode depth, spatial layout, and multi-view consistency \cite{wang2026vggt,lin2025depth}; reconstructive latents, such as VAE features \cite{podell2024sdxl,wan2025wan}, which retain fine-grained appearance and motion information and can be decoded back into pixels; and predictive latents, which are explicitly trained to capture temporally persistent and forecastable visual structure \cite{mur2026v}. We similarly instantiate and compare representatives from all four families within the same framework \cite{lyu2026lda}. In particular, it reveals which visual information is most useful for world-action modeling, rather than assuming that representations optimized for recognition or reconstruction are necessarily optimal for embodied control, as most existing papers do.

\textbf{World-Action Modeling Objectives.} Let $o_t$ denote the current observation, $\ell$ the task instruction, $a_{t+1:t+k}$ a future action chunk, and $z_{t+1:t+k}$ the corresponding future visual latents. We investigate four objectives: behavior cloning (BC), $p(a_{t+1:t+k} \mid o_t, \ell)$, as in standard VLA models; inverse dynamics modeling (IDM), $p(a_{t+1:t+k} \mid o_t, z_{t+1:t+k}, \ell)$; forward dynamics modeling (FDM), $p(z_{t+1:t+k} \mid o_t, a_{t+1:t+k}, \ell)$; and video generation (VG), $p(z_{t+1:t+k} \mid o_t, \ell)$. We instantiate all four objectives within the same framework \cite{lyu2026lda} using task-specific conditioning while holding all other configurations fixed. We compare BC-only training (representing the standard VLA setting), BC paired with each auxiliary objective, leave-one-objective-out variants, and joint training over all objectives. This controlled study explores whether these objectives provide complementary supervision or introduce competing learning signals.

% The training objective determines which conditional relationships a WAM learns and how visual-dynamics supervision contributes to action generation.

% whether policy improvements arise from action-conditioned future prediction, future-conditioned action inference, action-agnostic visual forecasting, or general multi-task regularization, while revealing 

\section{Main Findings and Results}\label{exp_setup}

\subsection{Video-Action Causality}\label{sec:vac}

\begin{figure*}[t]
\centering
  % \vspace*{-0.2in}
\begin{tikzpicture}[inner sep=0pt, outer sep=0pt]
  \node[anchor=south west] (fnC) at (0in,0in)
    {\includegraphics[
      height=1.4in,
      clip=true,
      trim=0in 0in 0in 0in
    ]{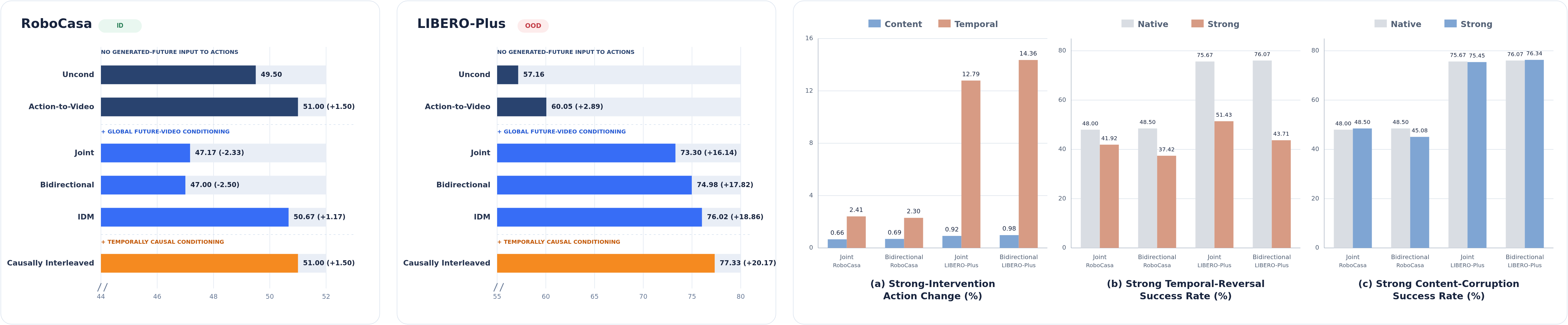}};
\end{tikzpicture}
\caption{Right: effects of generated-future interventions and policy performance. (a) Action-prediction change under strong content corruption and temporal reversal. (b) Native and strong temporal-reversal performance. (c) Native and strong content-corruption performance. Left: success rates (\%) on RoboCasa-GR1 and LIBERO-Plus.}
  % \vspace*{-0.2in}
\label{fig:vac_results}
\end{figure*}

\noindent\textbf{Does the generated future video influence action generation, or serve only as an auxiliary training signal?} To answer this question, four variants from Figure~\ref{fig:vac_results} (left) form two matched structural contrasts for the route of interest: Joint versus Uncond and Bidirectional versus Action-to-Video.  In each pair, the relevant difference is whether action tokens can read the generated future video.  IDM and Causally Interleaved remain useful policy variants in the overall comparison, but do not provide matched controls for this question. On the full LIBERO-Plus OOD benchmark, the route-enabled variants outperform their matched controls by 16.14\% and 14.93\%, whereas the same route-enabled variants underperform their controls by 2.33\% and 4.00\% on ID RoboCasa-GR1. Thus, future-video access is associated with improved OOD performance but not with ID gains. Since these comparisons involve separately trained policies, they cannot seriously establish whether generated futures causally affect action generation at inference time.

We therefore intervene directly on the generated future latents within the route-enabled Joint and Bidirectional policies, keeping other factors fixed. To distinguish reliance on exact future content from reliance on temporal organization, we use two complementary interventions: content corruption mixes each future slot with fixed noise at strengths 0.10, 0.25, and 0.50 while preserving its per-channel spatial mean and variance; temporal reversal swaps the two future slots during the final 25\%, 50\%, or 100\% of denoising. Both interventions leave the observed current-frame slot unchanged. We measure the resulting changes in action predictions and rollout success.

\begin{findingbox}{Takeaway 1.1: Generated futures affect actions mainly through temporal structure}
Temporal disruption has much larger effects under OOD shifts, while policies largely tolerate content corruption.
\end{findingbox}

As shown in Figure~\ref{fig:vac_results} (right), both ID and OOD evaluations show the same causal pattern. At the strongest level, content corruption changes action predictions by less than 1\% and produces no substantial change in success rate. In contrast, reversing the temporal order of the future latents changes actions by 2.30-2.41\% and reduces success by 6.08-11.08\% on RoboCasa-GR1, and changes actions more drastically by 12.79-14.36\% and reduces success by 24.24-32.37\% under OOD evaluation, spanning nearly all tasks and perceptual shifts. Policies, therefore, rely primarily on the temporal organization of generated futures, not their exact content. Results across all content-corruption and temporal-reversal strengths are provided in the appendix. Together, the structural comparisons and within-policy interventions show that generated futures are more than an auxiliary learning signal. Their temporal organization causally mediates action generation at inference time, but the resulting performance contribution is strongest under perceptual shift and does not imply a universal ID benefit.

\begin{findingbox}{Takeaway 1.2: Causal video generation appears to be the primary temporal advantage}
Causal video generation is a more important temporal component than temporal causality between video and action tokens.
\end{findingbox}

We next explore temporal structure at the architectural level. Among the six main variants, Causally Interleaved is the only one with a strictly temporal, framewise-interleaved multimodal attention graph, and it achieves the highest LIBERO-Plus success rate at 77.33\%. Together with the Takeaway~1.1, this ranking suggests that strict temporal organization may help explain its advantage. 

To further explore this design, we test a variant of Causally Interleaved, named Video-Causal Global, which
preserves causal video generation but allows every action group to read the complete temporally indexed video horizon. Video-Causal Global achieves a 79.84\% LIBERO-Plus success rate, 2.51\% above Causally
Interleaved. This improvement suggests that the advantage is more closely tied to frame-causal video generation than to enforcing strict temporal causality between video and action tokens.

% this session is reversed for lingbot causal!!!!!!!!!

\subsection{Latent World Representation}

% \textcolor{red}{TODO: (1) refine Figure 4 to make style consistent. (2) refine section B text. (3) add a compact conclusion in introduction section. (4) refine figure 1. check the mode of QWen-VL}

\begin{figure*}[t]
\centering
  % \vspace*{-0.2in}
\begin{tikzpicture}[inner sep=0pt, outer sep=0pt]
  \node[anchor=south west] (fnC) at (0in,0in)
    {\includegraphics[
      width=\linewidth,
      keepaspectratio
    ]{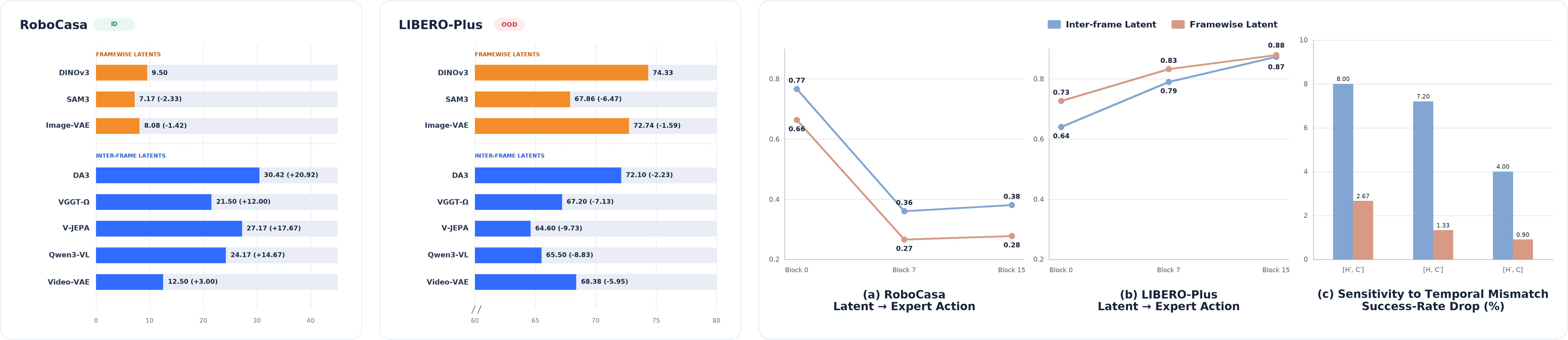}};
\end{tikzpicture}
\caption{Right: effects of temporal structure on action decodability, temporal sensitivity, and policy performance. (a, b) Held-out expert-action probing results. (c) ID performance degradation under temporal mismatch. Left: success rates (\%) on RoboCasa-GR1 and LIBERO-Plus.}
  % \vspace*{-0.2in}
\label{fig:latent_space_eval}
\end{figure*}

\noindent\textbf{Which property of latent world representations most strongly shapes WAM performance?}
We compare eight visual latent representations from four families: semantic, geometric, reconstructive, and predictive. Across these families, performance separates most clearly according to whether the visual latent itself encodes cross-frame temporal relations. DA3, VGGT-$\Omega$, V-JEPA, Qwen3-VL (video mode), and Video-VAE jointly encode multiple frames and therefore produce inter-frame latents. In contrast, DINOv3, SAM3, and Image-VAE encode each frame independently and produce framewise latents without explicit temporal relations. For framewise latents, temporal integration occurs only later in the policy trunk.

\begin{findingbox}{Takeaway 2.1: Temporal structure in visual latents shapes performance across distributions}
Inter-frame latents improve ID control, whereas framewise latents are favored under OOD.
\end{findingbox}

As shown in Figure~\ref{fig:latent_space_eval} (left), inter-frame latents outperform framewise latents on RoboCasa-GR1, but the ordering reverses on LIBERO-Plus, with the largest margins under sensor noise and camera-viewpoint shifts. Under our setup, this reversal suggests that inter-frame latents exploit temporal patterns that are effective on familiar trajectories but become less reliable when these patterns shift. Framewise latents instead preserve frame-local visual evidence and defer temporal integration to the downstream policy, potentially enabling greater adaptation under OOD shifts.

\noindent\textbf{Why does temporal structure produce opposite ID and OOD behavior?}
The performance reversal suggests that inter-frame and framewise latents may provide different action priors to the policy. To test this hypothesis, we examine whether the latent group that performs better in each setting also produces hidden states that are more directly aligned with expert actions. We fit linear probes that predict expert actions from hidden states at Blocks 0, 7, and 15. Higher held-out $R^2$ indicates that expert actions are more linearly recoverable from the corresponding hidden states. As shown in Figure~\ref{fig:latent_space_eval} (a) and (b), inter-frame latents are more action-decodable at every probed block on RoboCasa-GR1. On LIBERO-Plus, framewise latents instead lead at every block, although the difference becomes nearly negligible at the final block. This ordering mirrors the performance split, while the narrowing OOD gap suggests that deeper policy layers progressively reduce the initial difference between the two latent groups.

% The probes are trained on the training trajectories and evaluated on unseen held-out trajectories. 

The probing results show that action decodability follows the same distribution-dependent split as policy performance, but do not explain why inter-frame latents become less reliable under distribution shift. We hypothesize that their ID advantage depends on fixed temporal relations from the frozen visual encoder. To test this prediction, we perturb an ID input [H, C], where H and C denote the historical and current frames, by replacing either or both frames with earlier observations from the same trajectory. This produces [H', C], [H, C'], and [H', C'].

\begin{findingbox}{Takeaway 2.2: Pre-encoded inter-frame relations introduce fragile temporal dependencies}
Inter-frame latents degrade more when temporal relations are disrupted, whereas framewise latents remain more robust.
\end{findingbox}

As shown in Figure~\ref{fig:latent_space_eval} (c), across all three perturbations, policies using inter-frame latents degrade substantially more than those using framewise latents. Together with the probing results, this sensitivity suggests that pre-encoded temporal relations make expert actions more linearly decodable on familiar trajectories but introduce fragile dependencies. By deferring temporal modeling to the policy, framewise latents yield smaller ID gains but greater robustness under distribution shift. 

% when temporal structure is disrupted

% ; replacing both frames causes the largest degradation

% Together with Section~\ref{sec:vac}, these findings suggest that temporal structure is most beneficial when captured through world–action modeling, whereas fixed temporal relations pre-encoded in latent representations can introduce brittle dependencies under distribution shift.

% As shown in Figure~\ref{fig:latent_space_eval} (c), across all three perturbations, policies using inter-frame latents degrade substantially more than those using framewise latents. Together with the probing results, this sensitivity reveals a trade-off in explicitly encoding temporal relations in the latent representation. Inter-frame latents make expert actions more linearly decodable on familiar trajectories, suggesting that they expose task-relevant temporal structure more directly to the policy. However, this advantage comes with stronger dependencies on the particular temporal relations observed during training: when the input sequence is perturbed or shifted from the training distribution, these pre-encoded relations become unreliable and lead to substantially larger performance degradation. In contrast, framewise latents preserve less temporal structure at the representation level and therefore provide smaller in-distribution gains, but allow the policy to infer temporal relations from the current inputs rather than relying on fixed relationships embedded by the encoder. This results in greater robustness under distribution shift.

Together with Section~\ref{sec:vac}, these findings suggest that temporal structure is most beneficial when captured through world--action modeling rather than imposed directly by the latent representation. Modeling temporal relations within the policy allows them to remain conditioned on the current context, whereas fixed temporal relations pre-encoded in inter-frame latents can become brittle when those relations change at test time. Thus, while temporal information itself is valuable, where and how it is modeled plays an important role in determining whether it improves generalization or instead introduces fragile dependencies.

\subsection{World-Action Modeling Objectives}

\begin{figure*}[t]
\centering
  % \vspace*{-0.2in}
\begin{tikzpicture}[inner sep=0pt, outer sep=0pt]
  \node[anchor=south west] (fnC) at (0in,0in)
    {\includegraphics[
      height=1.25in,
      clip=true,
      trim=0in 0in 0in 0in
    ]{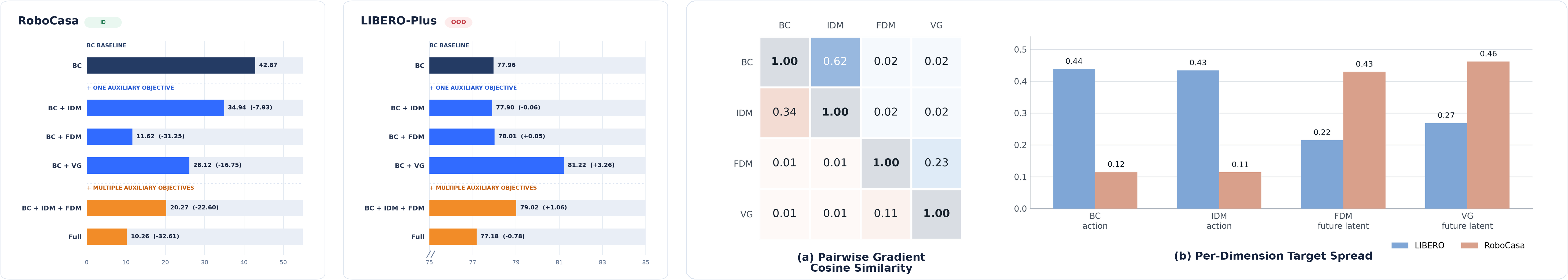}};
\end{tikzpicture}
\caption{Right: objective gradient alignment, target ambiguity, and policy performance. (a) Pairwise gradient alignment among objectives, measured by cosine similarity; the upper triangle reports LIBERO and the lower triangle reports RoboCasa-GR1. (b) Per-dimension target spread measures local variation in action targets (BC, IDM) or future latent targets (FDM, VG). Left: success rates (\%) on RoboCasa-GR1 and LIBERO-Plus.}
  % \vspace*{-0.2in}
\label{fig:mol_results}
\end{figure*}

\noindent\textbf{How does each auxiliary objective contribute to policy performance?}
We first examine how different auxiliary world-modeling objectives affect policy performance across ID and OOD settings.

% \begin{findingbox}{Takeaway 3.1: Auxiliary objectives do not improve in-distribution performance}
% BC-only performs best in distribution, while auxiliary objectives divert learning capacity.
% \end{findingbox}

\begin{findingbox}{Takeaway 3.1: Auxiliary objectives demonstrate distribution-dependent effects}
Auxiliary objectives hurt ID performance, but future-forecasting objectives, especially video generation, improve OOD robustness.
\end{findingbox}

% We focus on RoboCasa-GR1 because standard LIBERO is nearly saturated and provides limited separation among objective variants. All variants are optimized using the same training budget. 

As shown in Figure~\ref{fig:mol_results} (left), BC-only achieves the highest success rate on RoboCasa-GR1, while all auxiliary configurations perform worse. These results suggest that WAM effectively reduces to a standard VLA under ID setting, with auxiliary objectives diverting capacity from policy learning.  
IDM degrades performance less than the others because its gradient is more aligned with the BC objective as shown in Figure~\ref{fig:mol_results} (a). This stronger alignment allows IDM to better preserve the action-learning direction, even though it does not improve performance.

% \begin{findingbox}{Takeaway 3.2: Future forecasting provides more effective OOD regularization}
% Future-forecasting objectives, particularly video generation, improve robustness to perceptual shifts.
% \end{findingbox}

On LIBERO-Plus, VG improves BC-only OOD performance from 77.96\% to 81.22\%, whereas FDM provides only a marginal gain, and IDM slightly degrades performance. As shown in Figure~\ref{fig:mol_results} (a), IDM produces a strongly BC-aligned gradient and therefore behaves like an additional action-learning signal. FDM and VG instead provide nearly BC-orthogonal regularization whose benefits emerge primarily under perceptual shifts. For example, VG improves camera-viewpoint shift performance by 13.32\%, but does not improve performance under robot or layout shifts. Detailed statistics are provided in the appendix. 

 % weaker (much lower gradient norm), 

Although FDM and VG both forecast future states, VG provides stronger OOD generalization since it has a broader future target than FDM as shown in Figure~\ref{fig:mol_results} (b). This broader target encourages the representation to preserve richer visual-temporal information. More generally, dynamics objectives may be particularly sensitive to data coverage and training stage. With limited data or immature representations, dynamics objectives may minimize their losses using dataset-specific shortcuts, such as action-transition correlations, rather than learning transferable dynamics.

\noindent\textbf{Are the auxiliary objectives complementary or mutually interfering?}
We compare single-objective, jointly trained, and staged-training variants to determine whether the benefits of auxiliary objectives are additive.

\begin{findingbox}{Takeaway 3.2: Objective complementarity depends critically on the training schedule}
Naive joint training may cause objective interference, whereas staged training can improve performance.
\end{findingbox}

On LIBERO-Plus, BC+IDM+FDM modestly outperforms IDM and FDM individually. However, training all objectives jointly from the beginning reduces performance, which is 4.04\% below BC+VG. This indicates that naively imposing multiple objectives can interfere with the formation of transferable representations in our case. A staged schedule reverses this trend. Training with BC+VG during the first 80\% of optimization and introducing dynamics objectives only during the final 20\% achieves the highest 83.15\%. These results suggest that the objectives are not intrinsically incompatible. Early BC+VG training first establishes a stable policy and visual-temporal representation. The dynamics objectives can then act as weak late-stage regularizers without dominating representation formation or exploiting early training shortcuts. Importantly, the same staged strategy does not produce a corresponding ID improvement. This supports Takeaway~3.1. 

% to 77.18\%

\subsection{Real-Robot Data Experiments}

% \textbf{Setting.} We evaluate on DROID~\cite{khazatsky2024droid} using
% 24K/2K/5K language-labeled episodes for training, validation, and testing.
% Episodes are grouped by normalized task instruction, ensuring no exact
% instruction overlap across splits. We evaluate action prediction on one
% deterministic window per test episode and report normalized action MSE, L1,
% Accuracy@0.1, and Accuracy@0.5.

To examine whether our simulation findings transfer to real-robot data, we evaluate representative variants from each design axis on the held-out DROID test set. All variants are trained for 200K steps under matched settings. We report action MSE and L1 error, where lower values are better, together with Accuracy@0.1 and Accuracy@0.5, where higher values are better. Results are reported in Table~\ref{tab:real_robot}.

% To examine whether our simulation findings transfer to real-robot data, we evaluate representative variants from each design axis on the held-out DROID test set. All variants are trained for 200K steps under matched settings. We report action MSE and L1 error (lower is better) and Accuracy@0.1 and Accuracy@0.5 (higher is better).

\textbf{Video-Action Causality.} We compare Uncond with Causally Interleaved. Causally Interleaved consistently outperforms Uncond across all evaluated checkpoints from 10K to 200K steps and achieves better final-checkpoint performance on all four metrics. This supports our finding that generated futures provide more than auxiliary supervision: causally integrating them with action prediction improves generalization to unseen real-robot data. 

\textbf{Latent World Representation.} We compare the framewise DINOv3 representation with the inter-frame DA3 representation. DINOv3 performs better across all four metrics, providing further evidence that framewise latents can offer greater robustness on unseen real-robot data.

\textbf{World-Action Modeling Objectives.} We compare BC-only with BC+VG. BC+VG performs better across all four metrics, consistent with Takeaway~3.1 and further supporting the benefit of auxiliary video generation for generalization to unseen real-robot data.

\begin{table}[t]
\centering
\caption{Offline action-prediction results on the held-out DROID test set.}
\label{tab:droid_results}
  % \vspace*{-0.2in}
\resizebox{\linewidth}{!}{
\begin{tabular}{lcccc}
\toprule
\textbf{Variant} &
\textbf{MSE $\downarrow$} &
\textbf{L1 $\downarrow$} &
\textbf{Accuracy@0.1 $\uparrow$} &
\textbf{Accuracy@0.5 $\uparrow$} \\
\midrule

\multicolumn{5}{l}{\textbf{Video-Action Causality}} \\
Uncond
& 0.0730 & 0.1630 & 51.83\% & 93.87\% \\
Causally Interleaved
& \textbf{0.0666} & \textbf{0.1544}
& \textbf{53.78\%} & \textbf{94.52\%} \\
\midrule

\multicolumn{5}{l}{\textbf{Latent World Representation}} \\
DA3
& 0.0702 & 0.1612 & 52.40\% & 93.97\% \\
DINOv3
& \textbf{0.0684} & \textbf{0.1583}
& \textbf{53.08\%} & \textbf{94.05\%} \\
\midrule

\multicolumn{5}{l}{\textbf{World-Action Modeling Objective}} \\
BC-only
& 0.0769 & 0.1741 & 49.94\% & 91.84\% \\
BC+VG
& \textbf{0.0697} & \textbf{0.1656}
& \textbf{51.84\%} & \textbf{92.52\%} \\
\bottomrule
\end{tabular}
}
  % \vspace*{-0.2in}
\label{tab:real_robot}
\end{table}

\section{Discussion}

\textbf{Connections across the Three Design Choices.}
Taken together, our results identify temporal structure as a common factor connecting video-action causality, latent world representation, and training objectives, while showing that its benefit depends on where and how it is introduced. The causality study demonstrates that generated futures affect actions primarily through their temporal organization rather than their exact visual content, particularly under distribution shift. The representation study supports the importance of temporal information but also reveals an important qualification: inter-frame latents that directly entangle information across multiple frames can improve ID control while introducing temporal dependencies that become brittle under OOD shifts. These findings suggest that temporal structure is most useful when it is learned and integrated contextually within the WAM, rather than imposed as a fixed relation in the input representation.

The training objective study further supports this interpretation. Video generation provides a broad visual-temporal learning signal and improves OOD robustness, even though precise generated content has relatively little influence on action predictions in the intervention study. This suggests that video generation contributes both by organizing action generation temporally at inference time and by regularizing the learned representation during training. In contrast, dynamics objectives provide limited or inconsistent benefits and can interfere with policy learning when introduced too early. Their improvement under staged training indicates that the usefulness of world-modeling supervision depends not only on the objective itself, but also on when it is applied. Across all three design choices, a consistent distribution-dependent pattern emerges: design choices that encode stronger temporal priors can exploit familiar trajectories in distribution, whereas OOD robustness favors temporal structure that remains adaptable to the current context. Thus, world modeling is neither uniformly beneficial nor harmful; its value depends jointly on the causal pathway, representation, training schedule, and target evaluation distribution.

\textbf{Limitations and Future Work.}
Our conclusions are based primarily on controlled, moderate-scale experiments and should therefore be interpreted within the evaluated model families, datasets, and training budgets. This scope reflects an intentional trade-off: scaling every configuration substantially would have prevented the broad and structurally controlled
comparison of causal structures, latent representations, and training objectives that is central to this study. Although our simulation findings are further supported by experiments on real-robot data, it remains possible that some effects change with substantially larger models and datasets. Future work should examine these scaling
effects by validating the most promising designs under larger training budgets, broader data distributions, and closed-loop real-robot evaluation.

% Consequently, some observed rankings or interactions may change with larger models, broader data coverage, longer training, or deployment on real robots.

\section{Conclusion}\label{conclusion}

In this work, we present a controlled study that disentangles popular design choices in existing WAMs and examines not only their empirical effects but also how and why they shape model behavior and performance. This study unfolds around three fundamental questions in building WAMs: (1) video–action causality, (2) latent world representation, and (3) world–action modeling objectives. Through structurally controlled experiments on RoboCasa-GR1, LIBERO, and LIBERO-Plus, together with validation on real-robot data from DROID, we find that generated futures influence action generation primarily through their temporal organization, whereas fixed temporal relations pre-encoded in latent representations can become brittle under distribution shift. We further show that auxiliary world-modeling objectives are not uniformly beneficial, with their effectiveness depending on the training schedule. Overall, our findings suggest that effective WAM design requires jointly considering causal structure, latent representation, and training objective, rather than treating world modeling as a universally beneficial component. We hope these results provide a systematic understanding of how core design choices shape world–action modeling and offer useful principles for the development of future WAM systems.

\bibliographystyle{IEEEtran}
 % \balance
\bibliography{root}

\newpage

\end{document}